# TRACKING SYSTEM TO AUTOMATE DATA COLLECTION OF MICROSCOPIC PEDESTRIAN TRAFFIC FLOW

Kardi TEKNOMO
Doctoral Student
Graduate School of Information Sciences
Tohoku University
Aobayama, Aoba-ku, Sendai
980-8579 Japan
Fax: + 81 22 217 7505
Email: kardi@plan.civil.tohoku.ac.jp

Yasushi TAKEYAMA
Associate Professor
Graduate School of Information Sciences
Tohoku University
Aobayama, Aoba-ku, Sendai
980-8579 Japan
Fax: + 81 22 217 7505
Email: takeyama@plan.civil.tohoku.ac.jp

Hajime INAMURA
Professor
Graduate School of Information Sciences
Tohoku University
Aobayama, Aoba-ku, Sendai
980-8579 Japan
Fax: + 81 22 217 7494
Email: inamura@plan.civil.tohoku.ac.jp

**Abstract:** To deal with many pedestrian data, automatic data collection is needed. This paper describes how to automate the microscopic pedestrian flow data collection from video files. The study is restricted only to pedestrians without considering vehicular - pedestrian interaction. Pedestrian tracking system consists of three sub-systems, which calculates the image processing, object tracking and traffic flow variables. The system receives input of stacks of images and parameters. The first sub-system performs Image Processing analysis while the second sub-system carries out the tracking of pedestrians by matching the features and tracing the pedestrian numbers frame by frame. The last sub-system deals with a NTXY database to calculate the pedestrian traffic-flow characteristic such as flow rate, speed and area module. Comparison with manual data collection method confirmed that the procedures described have significant potential to automate the data collection of both microscopic and macroscopic pedestrian flow variables.

**Key Words:** Microscopic, Pedestrian, Image Processing, Automation, Tracking

## 1. INTRODUCTION

To decide the appropriate standard and control of pedestrian facilities, pedestrian studies, which consist of data collection and analysis, need to be done. Technological advance of computer and video processing over a decade has changed pedestrian studies significantly. Progression of analysis has demanded better data collection and the progress in data collection method improves the analysis further towards a more detailed design. May (1990) suggests that traffic flow characteristics could be divided into two categories, microscopic level and macroscopic level. Figure 1 shows the historical events in pedestrian studies and the position of automatic microscopic pedestrian data collection in the pedestrian studies. Microscopic level involves individual units with traffic characteristics such as time and space headway and individual speed. Pedestrian macroscopic flow characteristic such as flow, average speed and area module is gathered for macroscopic analysis as suggested by Fruin (1971). Since the microscopic pedestrian level of analysis was not so developed until recently, macroscopic level of analysis was well accepted. Therefore, pedestrian-flow data collection of macroscopic level is well advanced compared to the microscopic one.

On the other hand, improvement of pedestrian analysis has been developed regarding individual movements of pedestrian. Henderson (1974) attempted to develop a model for microscopic pedestrian motion using a fluid dynamic model that was revised by Helbing (1992a). Since the numerical solution of the mathematical model is very difficult, simulation is favorable Microscopic Pedestrian Simulation Model (MPSM) is a computer simulation model of pedestrian movements where every pedestrian in the model is treated as an individual. There are many types of MPSM and most of them do not relate to each other. Gipps, P.G. *et al* (1985) and Okazaki (1979) have started the microscopic simulation using cost and benefit cell or magnetic force model. Helbing, D. et al (1999) developed the social force model, while Blue, V.J *et al* (2000) developed a cellular automata model for pedestrians. The use of microscopic pedestrian simulation for evacuation purposes was developed by Watts (1987), Lovas (1994) and Thompson, P.A *et al* (1995). They used a queuing network model to analyze the MPSM.

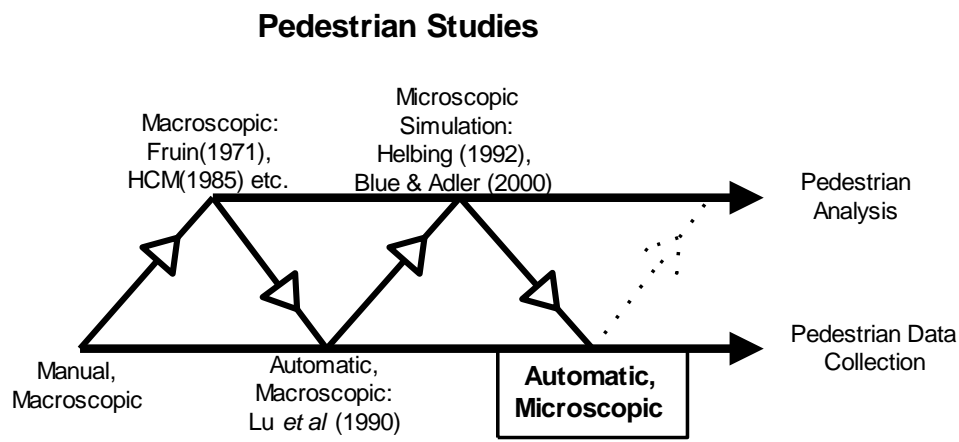

**Figure 1. Automatic Microscopic Data Collection in the Pedestrian Studies**

Most microscopic pedestrian simulations were not calibrated statistically and none of them have been calibrated using microscopic level data. It has no statistical guarantee that the parameters will work for general cases or even for a specific region. Such calibration was not possible without the ability to measure individual pedestrian-movement data. Since large amounts of data are needed for calibration purposes, automatic data-collection procedures should first be developed. Since the method of microscopic data-collection for pedestrian was not developed yet, research on automating data collection of individual data movements would be the appropriate step to use the microscopic level of analysis.

In accordance with pedestrian analysis, pedestrian data collection is developed to support the analysis. Progression of analysis has demanded better data collection and the progress in data collection method improves the analysis further toward a more detailed design. Typically, in earlier times, manual counting was performed using tally sheets or mechanical or electronic count boards to collect the volume and speed data for pedestrians. Pedestrian behavior studies are done by manual observation or video.

By the mid-90's, video and CCTV are increasingly popular as an "automatic" source of vehicular and pedestrian traffic data. Their advantage, to store the data in a videotape, which can be revisited to provide information on other aspects of the scene. Volume and speed data can be gathered separately for different times in the laboratory. Taping and filming provides an accurate and reliable means of recording volumes, as well as other data, but requires time-consuming data reduction in the office (ITE (1994)). The expense of reducing video data was very high because it must be done manually in the laboratory. Though vehicular automatic counting has been improved through pneumatic tube or inductance loops, the similar technology cannot be used to detect pedestrian.

The real automatic pedestrian counting device was developed as a research work of Lu, Y.J. (1990) using video camera. Macroscopic flow characteristic can be gathered automatically. The researchers however, limited themselves to special background and special treatment of the camera location. Mori, H. *et al* (1994) and Yasutomi, S. *et al* (1994) use cameras in side view and detect the rhythm of walking to discriminate pedestrian from other objects. Pedestrians moving alone can be tracked with very good accuracy. Tsuchikawa, M. *et al* (1995) used one line detection as the photo-beam technology developed to count the number of pedestrians passing that line using the top view camera. Significant advancement of pedestrian motion analysis was recently developed with a side view camera. Staufer, C. *et al* (2000) employed event detection and activities classification on the video camera for monitoring people activities (direction, coming and going). Ricquebourg, Y. *et al* (2000) exploited spatio-temporal XT slices to obtain trajectory patterns of a human walking. The method, however, has not been successful in detecting many pedestrians with occlusion cases. Haritaoglu, I. *et al* (2000) detected single and multiple people and monitoring their activities in an outdoor environment. It detects the people through their silhouettes and recognizes their activities with reasonable accuracy.

Though video surveillance of human motion has been developed, its utilization toward microscopic pedestrian data collection has never been employed yet. Research to combine these two fields of traffic engineering and computer vision is needed to develop a microscopic pedestrian data collection system. This paper describes how to automate the microscopic pedestrian flow data collection from a stack of images into pedestrian traffic flow characteristics. The pedestrian tracking system developed in this study has very important advantages toward the improvement of pedestrian traffic survey. The survey is a basic tool for designing and analyzing pedestrian facilities quantitatively.

The system developed by this study considers pedestrians in two-dimensional areas from a stack of images into pedestrian traffic flow variables. Pedestrians in stairs or elevators are not investigated. Mixed traffic between pedestrian and vehicular traffic is not examined either. The software was developed using Visual Basic 6.0 and it works under Microsoft Windows 95 operating systems or higher. The stack of images is presumed given by the hardware system and how to obtain it is out of the scope of this study.

## 2. PEDESTRIAN TRACKING SYSTEM

Figure 2 shows the structure of the Pedestrian Tracking System. The system receives input of stacks of images and parameters and produces pedestrian traffic flow characteristics as the output. Inside the system, there are three sub-systems. The first sub-system performs Image Processing analysis by making background images, removing the background from the image sequence to obtain objects in each frame, executing contour analysis and mask images toward each object to get object representation, segmentation and features. The output of the Image

Processing sub-system is a *features database*. The second sub-system carries out the tracking of pedestrians by matching the features and tracing the pedestrian numbers frame by frame. The output of this sub-system is a database called *NTXY database*, consisting of the pedestrian number, time and coordinate of each pedestrian at each time slice. The last sub-system deals with the NTXY database to calculate the pedestrian traffic-flow variables such as flow rate, speed and area module.

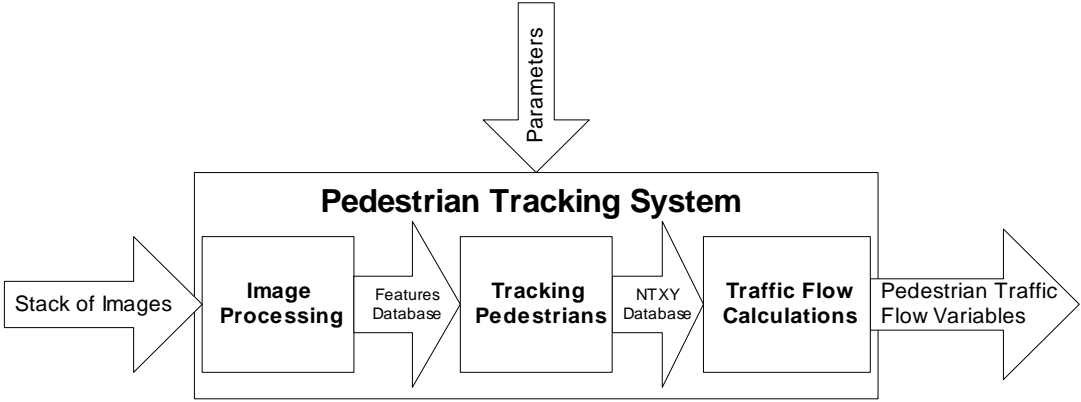

**Figure 2. Pedestrian Tracking System**

## 3. IMAGE PROCESSING

The image processing sub system consists of three modules: image segmentation, object detection and object descriptors. A video file can be seen as a collection of image sequence. Each image is called a slice or a frame. Fairhurst (1988) described an image as a two-dimensional function, where the value of the function $f(x,y)$ at spatial coordinate $(x,y)$ in the x-y plane defines a measure of light intensity of brightness (or gray level) at that point. Each spatial coordinate is called a pixel, representing a cell on a matrix containing an intensity level, which are digitized from the original continuous image function. Typically, a color image consists of three types of 256 gray level images, red, green and blue as seen in Figure 3. An image sequence or a stack is examined as a three-dimensional function $f(x,y,t)$ where $t$ is time. Both time and coordinates are in an integer set.

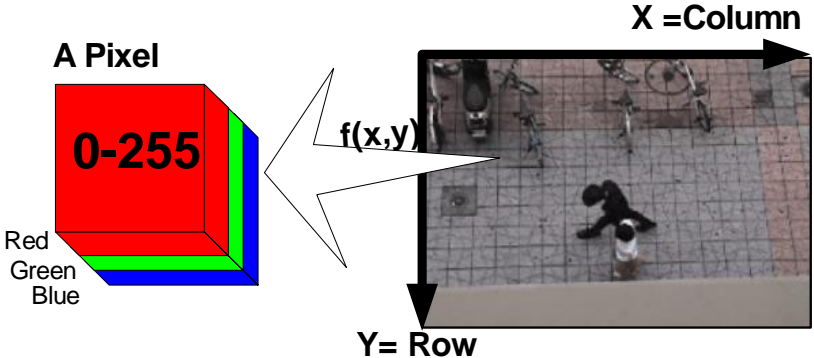

**Figure 3. Image representation and its coordinate system**

The target of the image processing procedure is to detect the location of pedestrians for each image. To detect the location of pedestrians, the pixels that picture the pedestrians must be first detected. If all pedestrians are using dark colored clothes and the background view or the environment is white, then the pixels of that pedestrian can easily be detected as dark colors. In reality, pedestrians may use any color of clothes and the environments are of different colors. To ease the detection of pedestrian pixels, the background view needs to be removed.

Given the image sequence, moving objects need to be separated from the static background. There are several methods to do the segmentation process. Background subtraction is one of the best tools for segmenting motion images from the static background. The algorithm is simple and fast compared to the optical flow method. Optical flow method needs a complex region grouping and it may be subjected to false grouping due to the assumption of uniform movement in grouping. Part of the pedestrian's body (e.g. feet and body), which has a different movement, will be grouped as different objects. Background subtraction also produces better segmentation compared to the image difference method. Image difference method leaves "ghosts" where the object was and leaves large regions of the objects undetected unless the object moves with significant motion in each frame.

If the background intensity level is $b(x,y)$ and the original-image intensity level is $f(x,y)$, then the object image, $g(x,y)$, is segmented by

$$g(x,y) = \begin{cases} 0 & \text{if } |f(x,y) - b(x,y)| < \theta \\ f(x,y) & \text{otherwise} \end{cases} \quad (1)$$

The object image has a white value (zero) at spatial coordinates (x, y) only if the intensity level difference between the original image and the background is small at the given coordinates, as determined by the threshold $\theta$. The threshold is determined by the variance of illumination in the short time (to reduce noise due to illumination change), normally taken as 10 to 25 out of 255 gray levels.

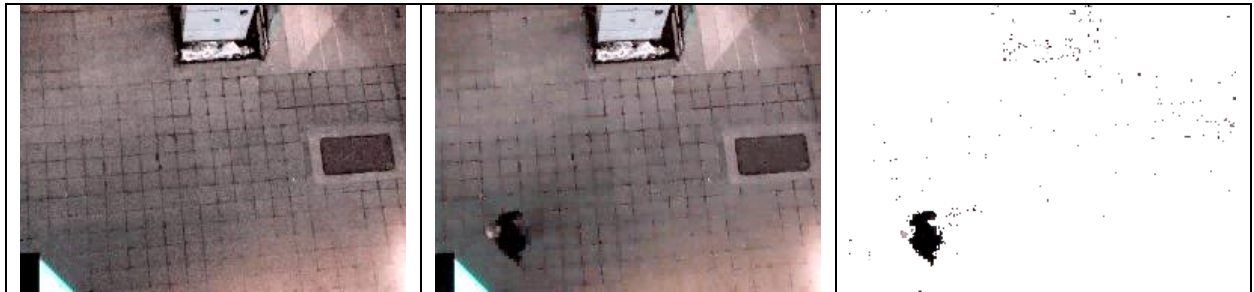

Figure 4. Background image (left), one slice in an image sequence (middle), result of subtracted images (right)

The above procedure can be performed provided that the background image shall be known or reconstructed by the existing image sequence. The background image can be obtained manually or automatically from using the existing algorithm (for example: Matsuyama, T. *et al* (1999)). The results from the background subtraction procedure, as shown in Figure 4, sometimes contain noise and disunity. Preprocessing of standard morphology closing and opening as suggested by Gonzalez, R.C. *et al* (1993) is performed before object detection to reduce the noise and increase the unity of the connected components of an object. After the moving objects are separated from the background, the pixels that contain the picture of pedestrians and other moving objects are detected in the binarized image. *Object Detection* stage isolate objects by representing them as regions of interest (ROI) or contour. The object detection based on 8 connected components identifies all the moving objects in the image. Contour is detected by the contour following method and chain codes. The image is scanned from the top left to bottom right. If a black pixel is found, the location of the pixel is stored and the neighboring pixels are searched as shown in Figure 5. If any of the eight directions of the neighbor pixels contains a black pixel, the black pixel is called connected. All the connected black pixels may produce a region that represents a moving object. Area threshold is determined to remove the noises. Each object detected is then bounded by a box to mark the object as detected.

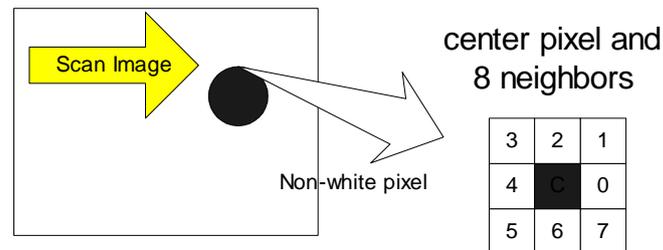

Figure 5. Object detection principle

Using the Region of Interest, some descriptors of objects can be calculated as *Object Features* or *Object Descriptors*. After an object in the binary image is detected, all pixels in the corresponding location of color image are captured. A bounding box can be drawn to mark detected object and object descriptors are calculated based on both the binary and color images. One object may have many descriptors to represent it. Pratt (1978) defines image feature or descriptor as a primitive characteristic or attribute of an image field. Contour extraction using 4 nearest neighbors based on Cabrelli C.A. *et al* (1990) is utilized to get one pixel line boundary of the black mask. After the objects are detected and contours are extracted, the features of each object are calculated.

A number of features are calculated based on the binary objects, color objects and the contour objects. Chain coding of the contour object gives the perimeter, object width and height. A binary object produces the area and centroid coordinate of the object. Statistical descriptors such as color mean, variance, skewness, kurtosis, as well as area, perimeters, compactness, center of gravity, and moments, etc. are also calculated. Each object detected is assumed one feature point that contains the aforementioned descriptors. Area of an object is the number of black pixels in the region bounded by the box. The coordinate of the moving object is calculated based on the center of gravity. Each black mask of binarized image has a corresponding pixel in the color image that can be measured as color object descriptors. The descriptors of each basic color, RGB, are measured and the averages of the three components represent the color descriptors. Mean color, color standard deviation and center of gravity of color is quantified as color descriptors.

For each slice-number, all object detected is put into a row of a database called *descriptor database*, as shown in Figure 6. The database consists of a field of slice object number, pedestrian number, slice number, coordinate locations (center of gravity) and all the other corresponding descriptors. Slice object number is an ordered number from 1 to maximum number of object detected on that slice. To each slice in a stack, a unique slice number is assigned from 1 to maximum number of frames in the stack. The field of pedestrian number in the descriptor database below is not calculated yet (i.e. equal to -1) because matching is not performed yet. After the tracking and recognition, this column is filled with unique pedestrian numbers.

| No Data | SlcObjNum | PedNum | sliceNum | cg_Area_X | cg_Area_Y | Area |
|---|---|---|---|---|---|---|
| 98 | 10 | -1 | 17 | 255.89 | 158.07 | 718 |
| 99 | 11 | -1 | 17 | 242.90 | 197.17 | 551 |
| 100 | 1 | -1 | 18 | 13.89 | 17.71 | 505 |
| 101 | 2 | -1 | 18 | 154.63 | 14.60 | 2690 |
| 102 | 3 | -1 | 18 | 5.95 | 77.11 | 769 |
| 103 | 4 | -1 | 18 | 46.13 | 82.56 | 375 |
| 104 | 5 | -1 | 18 | 215.67 | 79.76 | 324 |
| 105 | 6 | -1 | 18 | 128.79 | 90.36 | 711 |
| 106 | 7 | -1 | 18 | 57.30 | 139.87 | 1825 |
| 107 | 8 | -1 | 18 | 277.43 | 134.57 | 311 |
| 108 | 9 | -1 | 18 | 254.31 | 188.01 | 1957 |
| 109 | 1 | -1 | 19 | 13.92 | 17.10 | 508 |
| 110 | 2 | -1 | 19 | 160.78 | 11.88 | 2376 |

**Figure 6. Descriptor Database**

## 4. TRACKING PEDESTRIANS

Tracking an object as a point consists of a procedure to find the corresponding matching object based on similarity and closeness. Based on the feature database, the *voting descriptor* method is used to get the matched objects. Once the matched object is obtained, a frame based point tracking performs the tracing procedure. The intention of object matching in here is to find the most similar object between two slices. Firstly, let's consider two consecutive slices and each slice consists of several objects. Each object in each slice contains $\hbar$ number of descriptors.

An analogy of plurality voting is used to match the object similarity. The voter will be the descriptor and the candidates are all objects in slice s+1. The voter casts one vote or no votes for a given candidate. The calculation of voting follows these five rules:
1. Each descriptor may have one vote.
2. Each object (represented by the slice object number) in the next slice is a candidate.
3. If one or both descriptors are missing, it loses its vote.
4. If the absolute difference between the same descriptor on two consecutive slices is the minimum among all candidates, this candidate is winning for that descriptor. All other candidates lost the vote for that descriptor.
5. The matched candidate is the slice object number that has the maximum vote

Precisely, rule 1 to 4 can be expressed mathematically as follow. Let $\alpha'$ and $\alpha''$ are slice object number in slice $T$ and $T+1$ respectively, while $v$ and $w$ are maximum slice object number in slice $T$ and $T+1$. If $F_k(\alpha',T)$ and $F_k(\alpha'',T+1)$ are descriptor $k$-th of slice $T$ and $T+1$ respectively, then the vote is

$$\gamma^{\alpha'}(\alpha'',k) = \begin{cases} 1 & \text{if } \min_{\forall \alpha''} |F_k(\alpha',T) - F(\alpha'',T+1)| \\ 0 & \text{if } F_k(\alpha',T) \text{ or } F(\alpha'',T+1) = -1 \\ 0 & \text{otherwise} \end{cases} \qquad (2)$$

The first criterion in equation (2) is for the winning candidate, the second criterion covers the missing descriptors while the last criterion considers the losing candidate. After all descriptors

pass their vote, the winning candidate must be

$$\text{Match}(\alpha') = \{\alpha'' \Big| \max_{\forall \alpha''} \sum_k \gamma^{\alpha'}(\alpha'', k) \} \quad (3)$$

Slice object number $\alpha''$ is considered matched with slice object number $\alpha'$ if $\alpha''$ has a maximum vote. If $k^*$ are descriptors that are not missing, then a voting percentage is defined as

$$\vartheta = 100 \frac{\sum_k \gamma^{\alpha'}(\alpha'', k)}{\sum_k k^*} \quad (4)$$

A voting threshold is put on this voting percentage so that if the voting percentage is lower than the thresholds it means those objects are not similar. When all candidates in a slice have a voting percentage lower than the threshold, the matched candidate does not exist in the slice $T+1$. Searching the candidates from the next slices may end up with a matched object. The searching however, must be limited to a certain number of proceeding slices. An unmatched object, until a certain number of preceding slices, must be considered as a disappearing object or the object that has gone out of the scene. To reduce the number of unnecessary computation, a speed threshold is applied before the voting algorithm above. Only candidates that are inside a speed radius are considered valid candidates. All descriptors of all invalid candidates are considered as missing descriptors. In the case when the search for a matched candidate is done in slice $T+n$, then the searching speed radius should be multiplied by n to accommodate the motion of the object. Let $\theta_R$ be the speed threshold and $\vec{X}_{\alpha''}$ is the coordinate of object $\alpha'''$ slice $T+n$, then

$$\|\vec{X}_{\alpha''} - \vec{X}_{\alpha'}\| \geq n\theta_R \Rightarrow \forall_k F_k(\alpha''', T+n) = -1 \quad (5)$$

The speed threshold depends on the maximum speed of pedestrian and the scaling factors from the world to the images' coordinates.

In some cases, however, the winning candidate has been chosen previously. To avoid this case, the next candidate that has passed the voting threshold and obtained the maximum vote is considered as the winning candidate. The threshold of the voting percentage may determine the number of possible winning candidates. If the voting threshold is more than 50%, only one candidate will be chosen or none will be chosen. Similarly, if the voting threshold is set as 40%, there will be a maximum of 2 candidates when both candidates have the same similarity of 45%. Thus, the voting threshold depends on the number of allowable possible winning candidates. Therefore, matching is based on the similarity of descriptors and a matched object is an object that can pass these three criteria:
1. The speed between objects is smaller than speed threshold times counter of search. If it is found in n slices after, then the speed threshold will be larger (constant times n ).
2. The vote percentage is bigger than the voting threshold
3. The candidate must not have been chosen previously, except it is chosen to be the same object number.

A unique slice-object number is assigned to each object in the descriptor database. This slice-object number does not represent the real pedestrian number since it is only an ordered number for each frame. Each image of the pedestrian in each frame is represented by one point, which is the centroid of its area. Starting from the first frame, each point is numbered by a unique pedestrian-number. A simple tracing procedure is done to follow the movement-path coordinates of each pedestrian for every frame. The following steps are used for tracing the objects, given that the matching is done. To indicate an object, a pointer is used.

1. If the pointer is in the beginning of each slice, scan all objects in this slice. If there is no pedestrian number put pedestrian number based on maximum pedestrian number + 1.
2. For each object $\alpha'$ in a slice $T$, search for a match in the next slice or next $T+n$ slices.
3. If the matched object is found, stop searching and put the same pedestrian number in this new found object. Calculate the interpolation of coordinates and put the coordinates with the pedestrian number in the slice between $T$ and $T+n$. Put the pointer back to $\alpha'$. Go to step 5.
4. If not found in $T+n$ slices or until the end of stack, we assume the object has gone out of the scene. Put the pointer back to $\alpha'$.
5. If $\alpha'$ is not the last object of last slice, go to next object (i.e. new $\alpha'$). The slice number change according to the object. Go to step 1. If $\alpha'$ is in the last object of last slice, stop algorithm.

The result of this sub system is called the *NTXY database*. The database consists of four fields, which are pedestrian number, time and coordinate location. Pedestrian number, $\alpha$, is a unique number for each pedestrian and a new pedestrian number is given to a new person who enters the pedestrian trap. This number is useful to distinguish the data of a pedestrian from another. Only pedestrians in the pedestrian trap are recorded. Time recorded, $T$, represents the frame number. Because every two frames have a constant interval of $\Theta$ seconds, $T$ can be called as a discrete clock time. The coordinate location, X and Y, of each pedestrian is the real world coordinate of the pedestrian's image. Each row in the database represents a single observation point. One observation point is the position of a single pedestrian in one frame. The reference point of the coordinate system is arbitrary. For convention, in a common one or two-way traffic flow, the pedestrians are moving in the Y direction.

The images were taken at an angle from the pedestrian walkway, and cover a non-rectangular area. Due to lens distortion, the smallest error will be in the middle of the image and the highest error at the edge of the image. Pedestrian trap needs to be defined to reduce the lens distortion error and to ease the measurement. Though more sophisticated models exist (e.g. Tsai (1987)), a simple and satisfying model to convert Image Coordinates $(X_i, Y_i)$ to real world coordinates $(X_r, Y_r)$ was found using linear regression:

$$X_r = u + v.X_i + w.Y_i$$
$$Y_r = x + y.X_i + z.Y_i$$
(6)

## 5. PEDESTRIAN TRAFFIC FLOW CHARACTERISTICS

NTXY database that has been gathered consists of data of pedestrian movement. This huge data from the video is reduced into information that can be readily understood and interpreted which are traffic flow characteristics. In the NTXY database, the location of each pedestrian number N at time T is recorded as (X, Y). Using vector notation, the coordinate location of pedestrian N at time T is denoted as $\vec{X}_T$ because the pedestrian number N is obvious. The real time, when the pedestrian N is first and last recorded, is expressed by $T_i$ and $T_o$ respectively. Total number of observation, $\rho$, is the number of rows for pedestrian N in the database. The sampling time interval, $dt$, is a time gap between two recorded locations. Walking displacement of pedestrian N at time T is denoted by $\vec{\delta}_T = \vec{X}_{T+dt} - \vec{X}_T$ while its straight-line displacement is represented by $\vec{\Omega} = \vec{X}_{To} - \vec{X}_{Ti}$. The distance between pedestrian N with other pedestrians at time T is denoted by $\vec{d}_T$. If $\kappa$ represents the total number of pedestrian in the system during the time interval from $T_1$ until $T_2$, The pedestrian traffic flow characteristics can be formulated as shown in Table 1.

**Table 1. Pedestrian Traffic Flow Characteristics**

| Traffic Flow Variables | Formula | Eq. |
|---|---|---|
| Individual speed | $\upsilon = \dfrac{\sum_{t=Ti}^{To-1} \|\vec{X}_{t+dt} - \vec{X}_t\|}{To - Ti}$ | (7) |
| Headway | $h_T = \min(\|\vec{d}_T\|)$ | (8) |
| Flow rate | $q = \dfrac{\kappa}{T_2 - T_1}$ | (9) |
| Time mean speed | $TMS = \dfrac{\sum_N \upsilon_N}{\kappa}$ | (10) |
| Space mean speed | $SMS = \dfrac{L}{\dfrac{1}{\kappa}\sum_N \dfrac{L}{\upsilon_N}} = \dfrac{\kappa}{\sum_N \dfrac{1}{\upsilon_N}}$ | (11) |
| Area module | $M = \dfrac{\kappa}{A}$ | (12) |
| Moving Direction | $\vec{e} = \dfrac{\vec{\Omega}}{\|\vec{\Omega}\|}$ | (13) |

Average individual speed is the total walking distance of pedestrian N in the pedestrian trap divided by the total travel time in the pedestrian trap. Headway at time T is the minimum distance between pedestrian N with other pedestrians in the trap. Pedestrian traffic flow rate is the total number of pedestrians passing a line in the pedestrian trap over a certain time interval. Time mean speed is the average of individual speeds while the space mean speed is the average speed of all pedestrians occupying a given area of pedestrian trap over time interval $T_1$ until $T_2$. It is also equal to the ratio of the number of pedestrians and the harmonic summation of average individual speeds. The length of pedestrian trap is denoted by $L$. Area Module is the area of the pedestrian trap divided by the total number of pedestrian in the pedestrian trap over specified time interval. The moving direction of an individual pedestrian is determined by her first and last locations. The direction can be represented as a unit vector that connects the point when the pedestrian enters and egresses from the system. There are other indices that can be obtained from the NTXY database but the indices in Table 1 are the most common and important characteristics for pedestrian analysis.

## 6. DISCUSSION

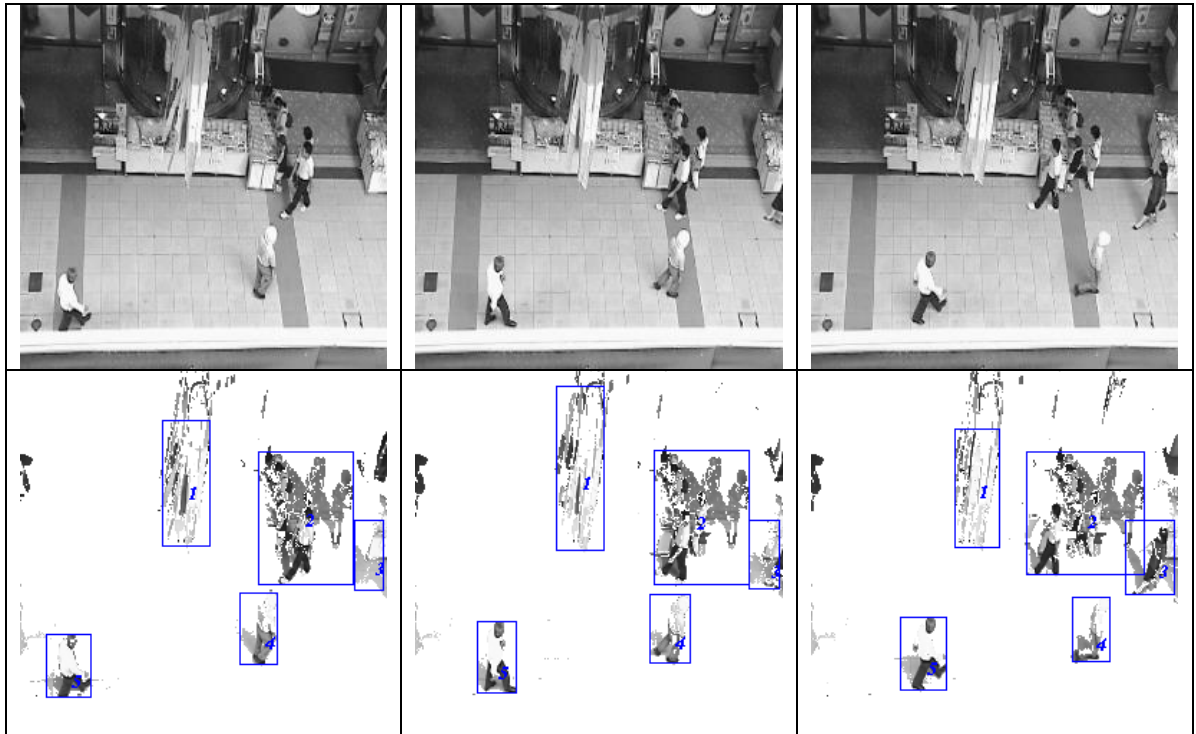

**Figure 7. Original images and result of Tracking**

To be able to examine the soundness of the algorithm proposed, several experiments were performed. The algorithm was implemented on real-world pedestrian traffic to demonstrate its working performance. Comparing the results of automatic processing with manual processing revealed that the algorithm was working sufficiently. For each image sequence, the object number is then compared with the image sequence according to its coordinate location. When false detection happens, on which features contribute to the falsely matched were investigated and another trial set of thresholds was set. The experimental process is repeated until the falsely matched is minimized. Several image sequences are collected and tested. The image sequences range from fully controlled background view and simple experiments in the laboratory, to the outdoor uncontrolled scene of pedestrians.

Figure 7 shows one of the results of the outdoor scene in the pedestrian mall. The same pedestrian numbers are given to the corresponding pedestrian in different images. When pedestrians are moving alone and do not overlap with each other, the system can detect them correctly. Some paper decoration in the top of sidewalk is still falsely detected as pedestrians. Due to a bad choice of background image, however, some "ghost" persons are also be falsely detected.

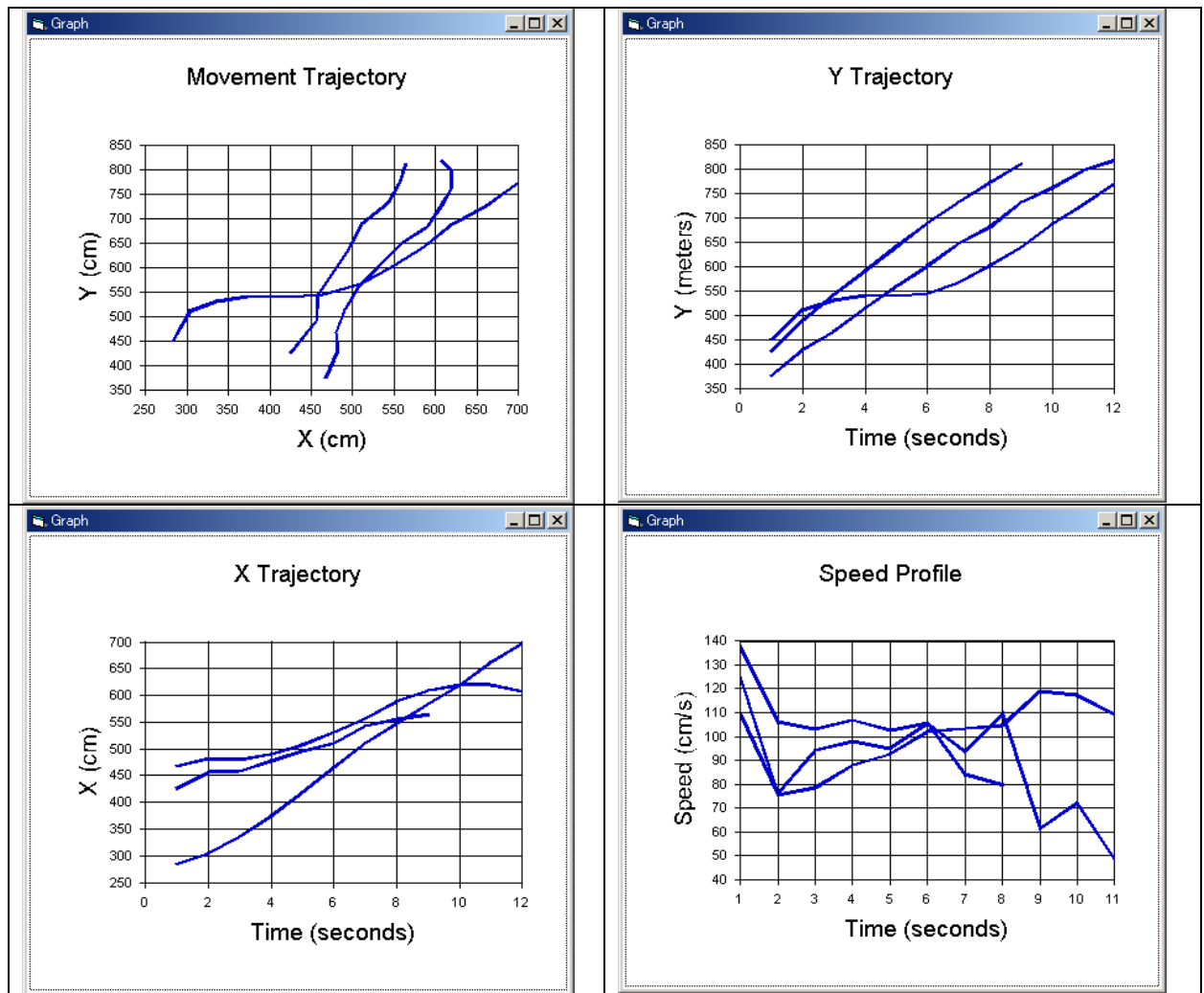
**Figure 8. Movement Trajectory in X and Y direction, and the speed profile**

Example of the result of the pedestrian tracking system is shown in Figure 8. The movements of three pedestrians are shown including their speed profiles. Since pedestrians are moving in two dimension, the trajectory can also be projected into X and Y direction. Comparing the speed profile and the movement trajectory, it reveals that the behavior of a not so smooth movement of two pedestrians in the movement trajectory can also be seen in the speed profile. The fluctuation in the speed profile represents the ragged behavior in the movement trajectory.

The experimental results revealed that the advance of the steps accumulates error of video processing. The error in background subtraction influences the error in the detection steps and descriptors calculations. Applying a set of appropriate thresholds however, may reduce these errors. Although each set of thresholds is determined by trial and error procedure, the following result of experiments can become a guideline. Figure 9 shows the relationship between background subtraction threshold, minimum area threshold and number of objects detected. As the threshold of background subtraction increase, the number of objects that are detected is reduced. Similarly, the number of objects detected is reduced when the threshold of minimum area is increased. The area of one pedestrian can be determined manually using the program, while threshold of background-subtraction is set to obtain the maximum number of objects detected for a specified minimum area threshold.

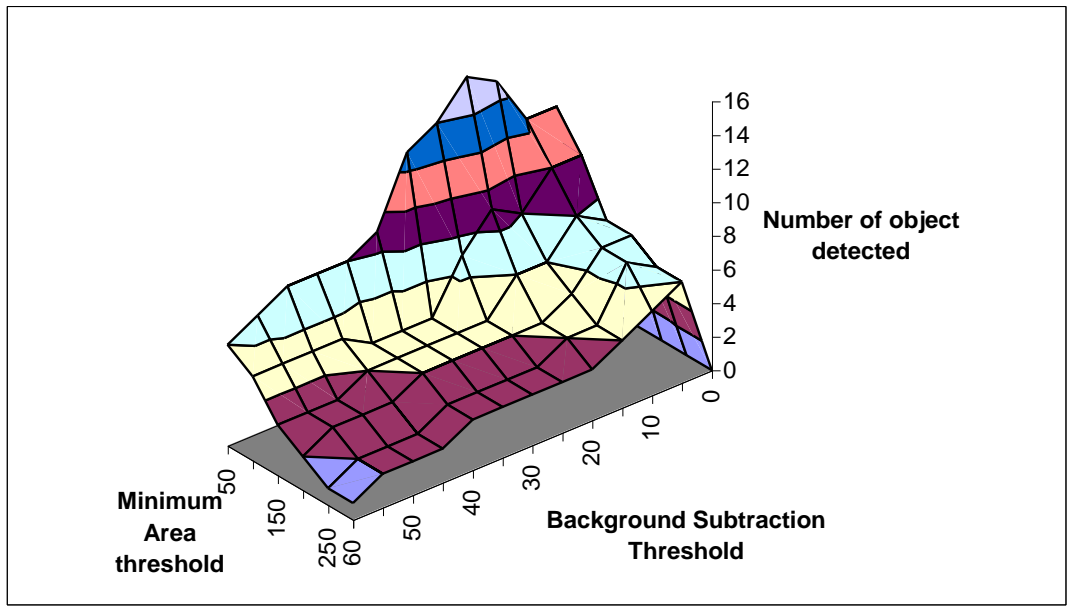

**Figure 9. Relationship of number of objects detected and its thresholds**

Tracking and matching steps are crucial since the highest error happens in this step due to partial occlusion problems. When projection of a 3D object toward a plane may produce a view of overlapping objects, an occlusion scene happens. The actual objects do not merge with each other, but the projection on the camera plane may display that view. Occlusion is a phenomenon wherein two or more objects may move too close to each other and is detected as one object. Later these objects may separate again as an individual object. A false positive (additional number of objects) happens merely due to partial occlusion. A merging event only is falsely traced as an object going out while diverging is falsely traced as a new coming object. A false negative happens when several pedestrians who are always together throughout the scene and because of merging event, is always detected as one object. The errors mostly happen due to matching and unrecoverable event by the changing of some thresholds.

Aside from those errors, the result of video processing was satisfactory since it is able to detect and track the movement of pedestrians who walk alone. It can detect the new coming and going out of pedestrians and full occlusion (merging and diverging) in short slices ($n \leq 3$) with very small error. The tracking was also performed correctly to detect new coming objects even after a long gap of no objects in the scene. For light pedestrian density, the video data collection produces excellent results. It was confirmed that the procedures described have significant potential to automate measurement of both microscopic and macroscopic pedestrian flow characteristics.

## 7. CONCLUSIONS

Technological advance of computer and video processing over a decade has changed pedestrian studies significantly. Progression of analysis has demanded better data collection and the progress in data collection method improves the analysis further toward a more detailed design. Recent trends of pedestrian studies have been dealing with microscopic level of analysis. Microscopic pedestrian tracking system has been described in this paper. The tracking system consists of three sub-systems, which calculates the image processing, object tracking and traffic flow variables. The system receives input of stacks of images and parameters. The first sub-system performs Image Processing analysis while the second sub-system carries out the tracking of pedestrians by matching the features and tracing the pedestrian numbers frame by frame. The last sub-system deals with the NTXY database to calculate the pedestrian traffic-flow characteristic such as flow rate, speed and area module. Comparison with manual data collection method reveals that the procedures described could deal with light pedestrian density with satisfactory result. It was confirmed that the procedures described have significant potential to automate the measurement of both microscopic and macroscopic pedestrian flow characteristics. Improvement of such system to deal with higher density pedestrian traffic and obtaining the fundamental traffic flow diagram is recommended for further study.